\documentclass[nohyperref]{article}

\usepackage[utf8]{inputenc}
\usepackage[lf,enc=t1]{berenis}

\usepackage{microtype}
\usepackage{graphicx}
\usepackage{subfigure}
\usepackage{booktabs} 
\usepackage{multirow}

\usepackage{algorithmicx}
\usepackage[ruled]{algorithm}
\usepackage{algpseudocode}
\usepackage{cancel}
\usepackage{makecell}
\usepackage{hyperref}

\def\our{HyperShot}

\usepackage[accepted]{icml2022}

\usepackage{amsmath}
\usepackage{amssymb}
\usepackage{mathtools}
\usepackage{amsthm}

\usepackage[capitalize,noabbrev]{cleveref}

\theoremstyle{plain}

\theoremstyle{definition}

\theoremstyle{remark}

\usepackage[textsize=tiny]{todonotes}

\icmltitlerunning{\our{}: Few-Shot Learning by Kernel HyperNetworks}

\begin{document}

\twocolumn[
\icmltitle{\our{}: Few-Shot Learning by Kernel HyperNetworks}



\icmlsetsymbol{equal}{*}

\begin{icmlauthorlist}

\icmlauthor{Marcin Sendera}{equal,uj}
\icmlauthor{Marcin Przewi\k{e}źlikowski}{equal,uj}
\icmlauthor{Konrad Karanowski}{pwr}\\
\icmlauthor{Maciej Zi\k{e}ba}{pwr,tooploox}
\icmlauthor{Jacek Tabor}{uj}
\icmlauthor{Przemysław Spurek}{uj}
\end{icmlauthorlist}

\icmlaffiliation{uj}{Faculty of Mathematics and Computer
Science, Jagiellonian University, Kraków, Poland}
\icmlaffiliation{pwr}{Wrocław University of Science and Technology, Wrocław, Poland}
\icmlaffiliation{tooploox}{Tooploox}

\icmlcorrespondingauthor{Marcin Sendera}{marcin.sendera@doctoral.uj.edu.pl}

\icmlkeywords{few-shot learning, meta-learning, kernels methods, hypernetworks}

\vskip 0.3in
]



\printAffiliationsAndNotice{\icmlEqualContribution} 

\begin{abstract}
Few-shot models aim at making predictions using a minimal number of labeled examples from a given task. The main challenge in this area is the \textit{one-shot} setting where only one element represents each class. We propose \textit{HyperShot} - the fusion of kernels and hypernetwork paradigm.
Compared to reference approaches that apply a gradient-based adjustment of the parameters, our model aims to switch the classification module parameters depending on the task's embedding. In practice, we utilize a hypernetwork, which takes the aggregated information from support data and returns the classifier's parameters handcrafted for the considered problem. 
Moreover, we introduce the kernel-based representation of the support examples delivered to hypernetwork to create the parameters of the classification module. 
Consequently, we rely on relations between embeddings of the support examples instead of direct feature values provided by the backbone models. 
Thanks to this approach, our model can adapt to highly different tasks. 
\end{abstract}

\section{Introduction}
\label{intro}

Current Artificial Intelligence techniques cannot rapidly generalize from a few examples. This common inability is because most deep neural networks must be trained on large-scale data. In contrast, humans can learn new tasks quickly by utilizing what they learned in the past. {\bf Few-shot learning} models try to fill this gap by \textit{learning how to learn} from a limited number of examples.
Few-shot learning is the problem of making predictions based on a small number of samples. The goal of few-shot learning is not to recognize a fixed set of labels but to learn how to quickly adapt to new tasks with a small amount of training data. After training, the model can classify new data using only a few training samples.


\begin{figure*}[t]
    \vskip 0.2in
    \begin{center}
    \centerline{\includegraphics[width=0.8\textwidth]{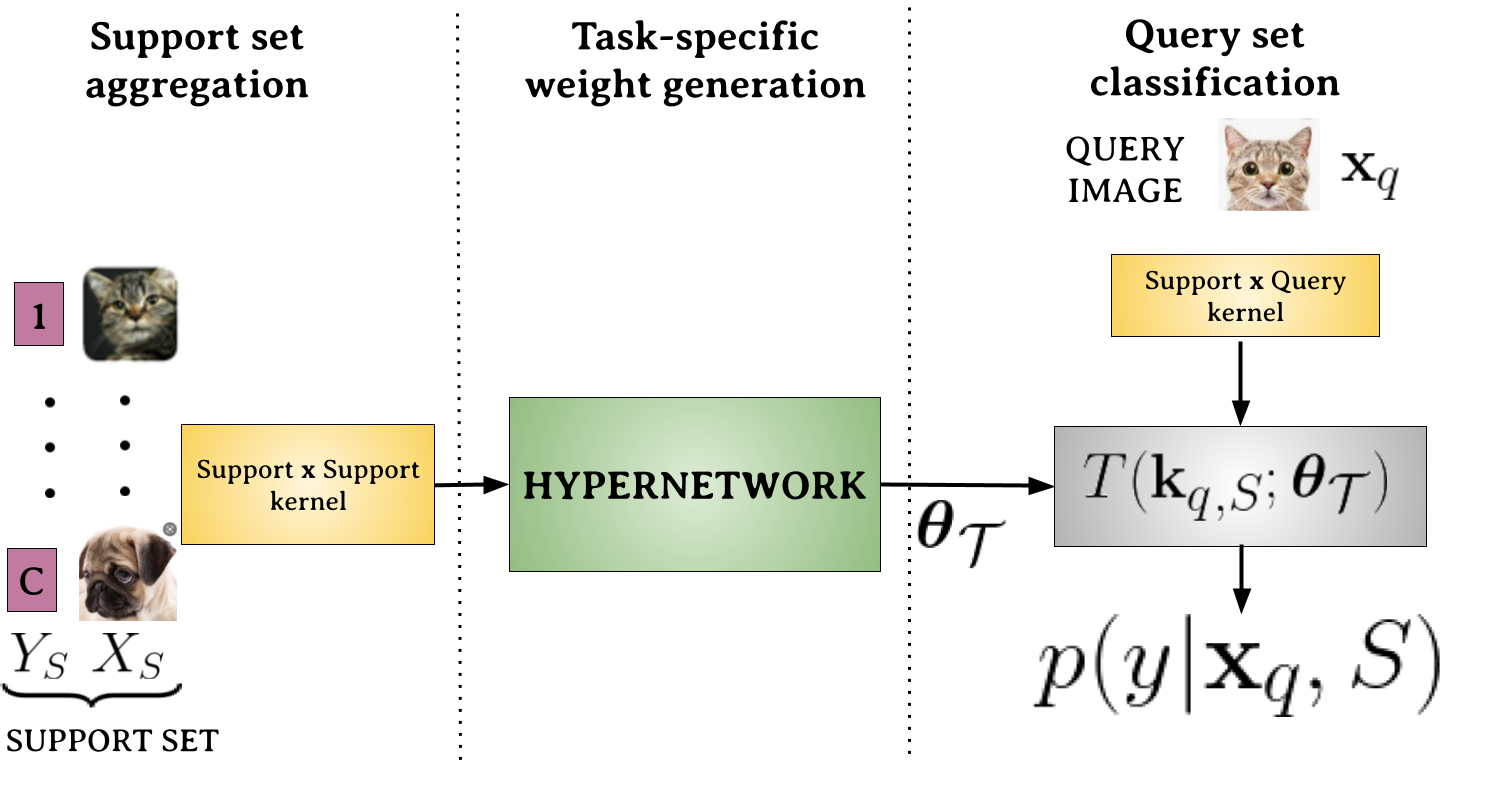}}
    \caption{The overview of \our{} architecture. We create the kernel matrix for features extracted from support examples, which is further processed by a hypernetwork. The role of the hypernetwork is to create the set of parameters for the considered task. The target network further uses the parameters to classify the query examples. }
    \label{fig:teaser}
    \end{center}
    \vskip -0.3in
\end{figure*}

The most popular approaches for few-shot learning use the meta-learning framework. We sample few-shot classification tasks from training samples belonging to the base classes and optimize the model to perform well on these tasks.
We work with $N$-way and $K$-shot tasks, where we have $N$ classes with $K$ support samples and $Q$ query samples in each category.
The goal is to classify these $N \times Q$ query samples into the $N$ classes based on the $N \times K$ support samples.

One of the most common approaches to Few-shot learning is Model-Agnostic Meta-Learning (MAML) \cite{finn2017model}, where the model is trained to adapt to new Few-shot learning tasks quickly. The main idea is to produce universal weights which can be rapidly updated to solve new small tasks. The model's main limitation is the complicated optimization procedure that uses an inner and outer loop, which can be seen as second-order optimization.
Moreover, such an optimization procedure only partially implements the \emph{learn how to learn} paradigm, which assumes the model learns some rules to adapt to the new tasks. The induction of such rules by MAML is limited by gradient-based optimization applied to adjust the parameters of the classification module to the new task. In practice, we can realize learning rules which are based on gradient optimization. New classification module parameters are obtained by gradient descent. 

This paper introduces a new strategy, which solves the above problem with restriction only to gradient-based decision rules. Our goal is to mimic the human learning process. First, we look at the entire support set and extract the information to distinguish objects in classes. Then, based on the relation between features, we create the decision rules.

We combine the Hypernetworks paradigm with kernel methods to realize such a scenario, see Fig.~\ref{fig:teaser}.
Hypernetworks, introduced in \cite{ha2016hypernetworks}, are defined as neural models that generate weights for a separate target network solving a specific task.  In our approach, the Hypernetwork aggregates the information from the support set and produces weights of the target model, classifying the query set.

Kernel methods realize the first part of the process. For each of the few-shot tasks, we extract the features from the support set through the backbone architecture and calculate kernel values between them. Then we use a Hypernetwork architecture -- a neural network that takes kernel representation and produces decision rules in the form of a classifier (target network) dedicated to the query set.  

Such a solution realizes the \emph{learn how to learn} paradigm. Hypernetwork, which produces the decision rules from kernel representation of the support set, is trained by gradient-based method and can be seen as a classical learning process. However, the weights of the target network (decision rules) can realize different strategies and lay in varying weight space parts. Our approach allows training a universal architecture model, which can be quickly updated to new tasks without any second-order procedure (inner and outer loop).

We perform an extensive experimental study of our approach by benchmarking it on various one-shot and few-shot image classification tasks. We find that \our{} demonstrates high accuracy in all tasks, performing comparably or better than the other recently proposed methods. Moreover, \our{} shows a strong ability to generalize, as evidenced by its performance on cross-domain classification tasks. 

The contributions of our work can be summarized as follows:
\begin{itemize}
    \item In this paper, we propose a model which realizes the \emph{learn how to learn} paradigm by modeling learning rules which are not based on gradient optimization and can produce completely different strategies.
    \item We propose a new approach to solve the few-shot learning problem by aggregating information from the support set and directly producing weights from the neural network dedicated to the query set.
    \item We propose \our{}, which uses the Hypernetwork paradigm to produce the weights dedicated for each task. 
\end{itemize}


\section{\our{}: Hypernetwork for few-shot learning}

In this section, we present our \our{} model for few-shot learning. 

\subsection{Background}

\paragraph{Few-shot learning}
The terminology describing the few-shot learning setup is dispersive due to the colliding definitions used in the literature. For a unified taxonomy, we refer the reader to \cite{chen2019closer,wang2020fewshotsurvey}. 
Here, we use the nomenclature derived from the meta-learning literature, which is the most prevalent at the time of writing.
Let: 
\begin{equation}
\mathcal{S} = \{ (\mathbf{x}_l, \mathbf{y}_l) \}_{l=1}^L
\end{equation}
be a support-set containing input-output pairs, with $L$ examples with the equal class distribution. In the \textit{one-shot} scenario, each class is represented by a single example, and $L=K$, where $K$ is the number of the considered classes in the given task. Whereas, for \textit{few-shot} scenarios, each class usually has from $2$ to $5$ representatives in the support set $\mathcal{S}$. Let:

\begin{equation}
\mathcal{Q} = \{ (\mathbf{x}_m, \mathbf{y}_m) \}_{m=1}^M
\end{equation}
be a query-set (sometimes referred to in the literature as a target-set), with $M$ examples, where $M$ is typically one order of magnitude greater than $K$. For ease of notation, the support and query sets are grouped in a task $\mathcal{T} = \{\mathcal{S}, \mathcal{Q} \}$. During the training stage, the models for few-shot applications are fed by randomly selected examples from training set $\mathcal{D} = \{\mathcal{T}_n\}^N_{n=1}$,  defined as a collection of such tasks. 

 During the inference stage, we consider task $\mathcal{T}_{*} = \{\mathcal{S}_{*}, \mathcal{X}_{*}\}$, where $\mathcal{S}_{*}$ is a support set with the known class values for a given task, and $\mathcal{X}_{*}$ is a set of query (unlabeled) inputs. The goal is to predict the class labels for query inputs $\mathbf{x} \in \mathcal{X}_*$, assuming support set $\mathcal{S}_{*}$ and using the model trained on $\mathcal{D}$.

\paragraph{Hypernetwork}

In the canonical work \cite{ha2016hypernetworks}, hyper-networks are defined as neural models that generate weights for a separate target network solving a specific task.
The authors aim to reduce the number of trainable parameters by designing a hyper-network with a smaller number of parameters than the target network. Making an analogy between hyper-networks and generative models, the authors of \cite{sheikh2017stochastic} use this
mechanism to generate a diverse set of target networks approximating the same function.

\subsection{\our{} - overview}

We introduce our {\our{}} - model that utilizes hypernetwork for few-shot problems. The main idea of the proposed approach is to predict the values of the parameters for a classification network that makes predictions on query images given the information extracted from support examples for a given task. Thanks to this approach, we can switch the classifier's parameters between completely different tasks based on the support set. The information about the current task is extracted from the support set using parametrized kernel function that operates on embedding space. Thanks to this approach, we use relations among the support examples instead of taking the direct values of the embedding values as an input to the hypernetwork. Consequently, this approach is robust to the embedding values for new tasks far from the feature regions observed during training. The classification of the query image is also performed using the kernel values calculated with respect to the support set.     

\begin{figure*}[t]
    \vskip 0.2in
    \begin{center}
    \centerline{\includegraphics[width=0.9\textwidth]{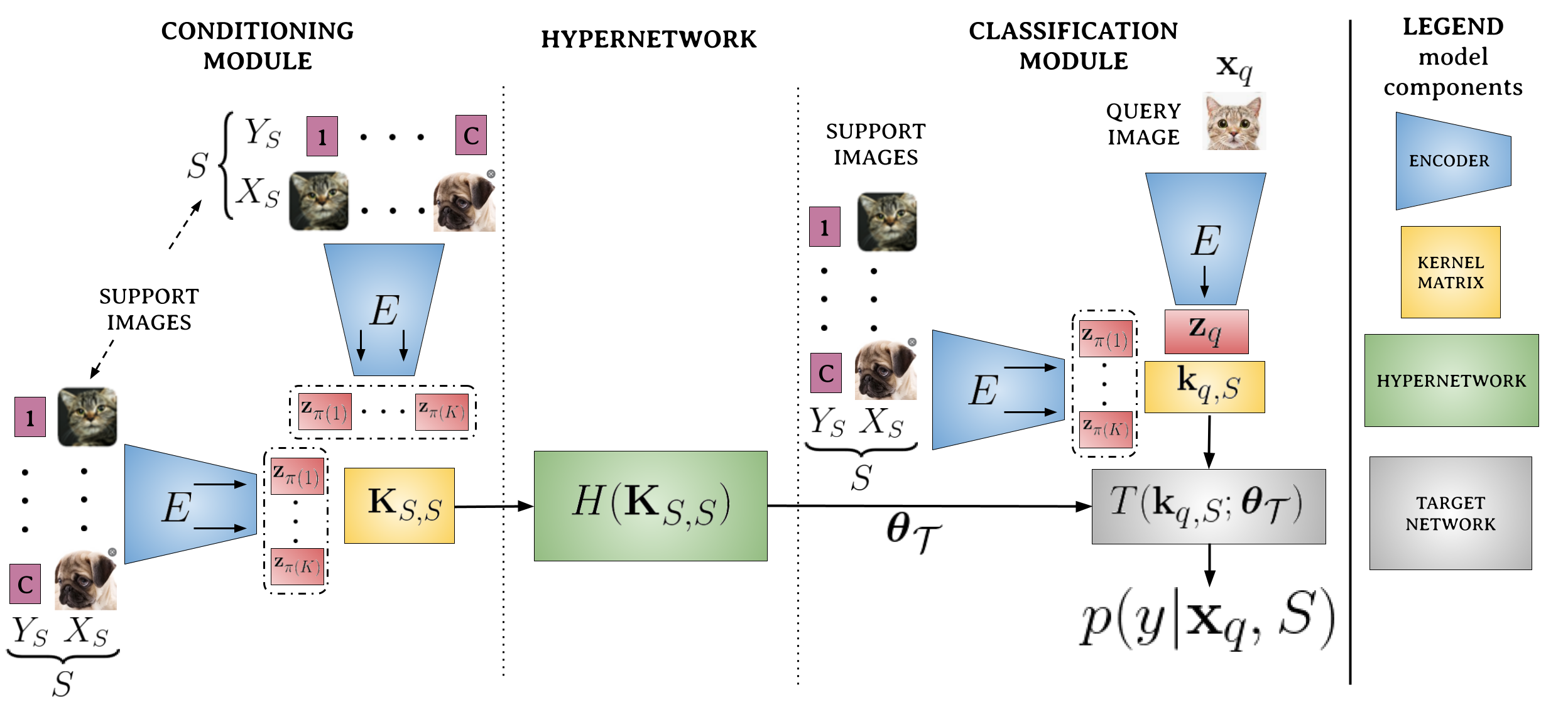}}
    \caption{The general architecture of \our{.} First, the examples from a support set are sorted according to the corresponding class labels and transformed by encoding network  $E(\cdot)$ to obtain the matrix of ordered embeddings of the support examples, $\mathbf{Z}_S$. The low-dimensional representations stored in $\mathbf{Z}_S$ are further used to compute kernel matrix $\mathbf{K}_{S,S}$. The values of the kernel matrix are passed to the hypernetwork $H(\cdot)$ that creates the parameters $\boldsymbol{\theta}_T$ for the target classification module $T(\cdot)$. The query image $\mathbf{x}$ is processed by encoder $E(\cdot)$, and the vector of kernel values $\mathbf{k}_{\mathbf{x},S}$ is calculated between query embedding $\mathbf{z}_{\mathbf{x}}$ and the corresponding representations of support examples, $\mathbf{Z}_S$. The kernel vector $\mathbf{k}_{\mathbf{x},S}$ is further passed to target model $T(\cdot)$ to obtain the probability distribution for the considered classes.   } 
    \label{fig:architecture}
    \end{center}
    \vskip -0.2in
\end{figure*}

The architecture of the \our{} is provided in Fig.~\ref{fig:architecture}. We aim to predict the class distribution $p(\mathbf{y}|S, \mathbf{x})$, given a query image $\mathbf{x}$ and set of support examples $S = \{ (\mathbf{x}_l, y_l) \}_{l=1}^K$. First, all images from the support set are grouped by their corresponding class values. Next, each of the images $\mathbf{x}_l$ from the support set is transformed using encoding network $E(\cdot)$, which creates low-dimensional representations of the images, $E(\mathbf{x}_l)=\mathbf{z}_l$. The constructed embeddings are sorted according to class labels and stored in the matrix $\mathbf{Z}_S=[\mathbf{z}_{\pi(1)}, \dots, \mathbf{z}_{\pi(K)}]^\mathrm{T}$, where $\pi(\cdot)$ is the bijective function, that satisfies $y_{\pi(l)} \leq y_{\pi(k)}$ for $l \leq k$. 

In the next step we calculate the kernel matrix $\mathbf{K}_{S, S}$, for vector pairs stored in rows of $\mathbf{Z}_S$. To achieve this, we use the parametrized kernel function $k(\cdot, \cdot)$, and calculate $k_{i,j}$ element of matrix $\mathbf{K}_{S, S}$ in the following way:

\begin{equation}
k_{i,j} = k(\mathbf{z}_{\pi(i)}, \mathbf{z}_{\pi(j)}).
\end{equation}

The kernel matrix $\mathbf{K}_{S, S}$ represents the extracted information about the relations between support examples for a given task. The matrix $\mathbf{K}_{S, S}$ is further reshaped to the vector format and delivered to the input of the hypernetwork $H(\cdot)$. The role of the hypernetwork is to provide the parameters $\boldsymbol \theta_T$ of target model $T(\cdot)$ responsible for the classification of the query object. Thanks to that approach, we can switch between the parameters for entirely different tasks without moving via the gradient-controlled trajectory, like in some reference approaches like MAML. 

The query image $\mathbf{x}$ is classified in the following manner. First, the input image is transformed to low-dimensional feature representation $z_{\mathbf{x}}$ by encoder $E(\mathbf{x})$. Further, the kernel vector $\mathbf{k}_{\mathbf{x},S}$ between the query embedding and sorted support vectors $\mathbf{Z}_S$ is calculated in the following way:

\begin{equation}
\mathbf{k}_{\mathbf{x}, S} = [k(\mathbf{z}_{\mathbf{x}}, \mathbf{z}_{\pi (1)}),\dots,k(\mathbf{z}_{\mathbf{x}}, \mathbf{z}_{\pi (K)})]^{\mathrm{T}}.  
\end{equation}

The vector $\mathbf{k}_{\mathbf{x}, S}$ is further provided on the input of target model $T(\cdot)$ that is using the parameters $\boldsymbol \theta_{T}$ returned by hypernetwork $H(\cdot)$. The target model returns the probability distribution $p(\mathbf{y}|S, \mathbf{x})$ for each class considered in the task. 

\begin{figure}[t]
    \begin{center}
    \centerline{\includegraphics[width=0.47\textwidth]{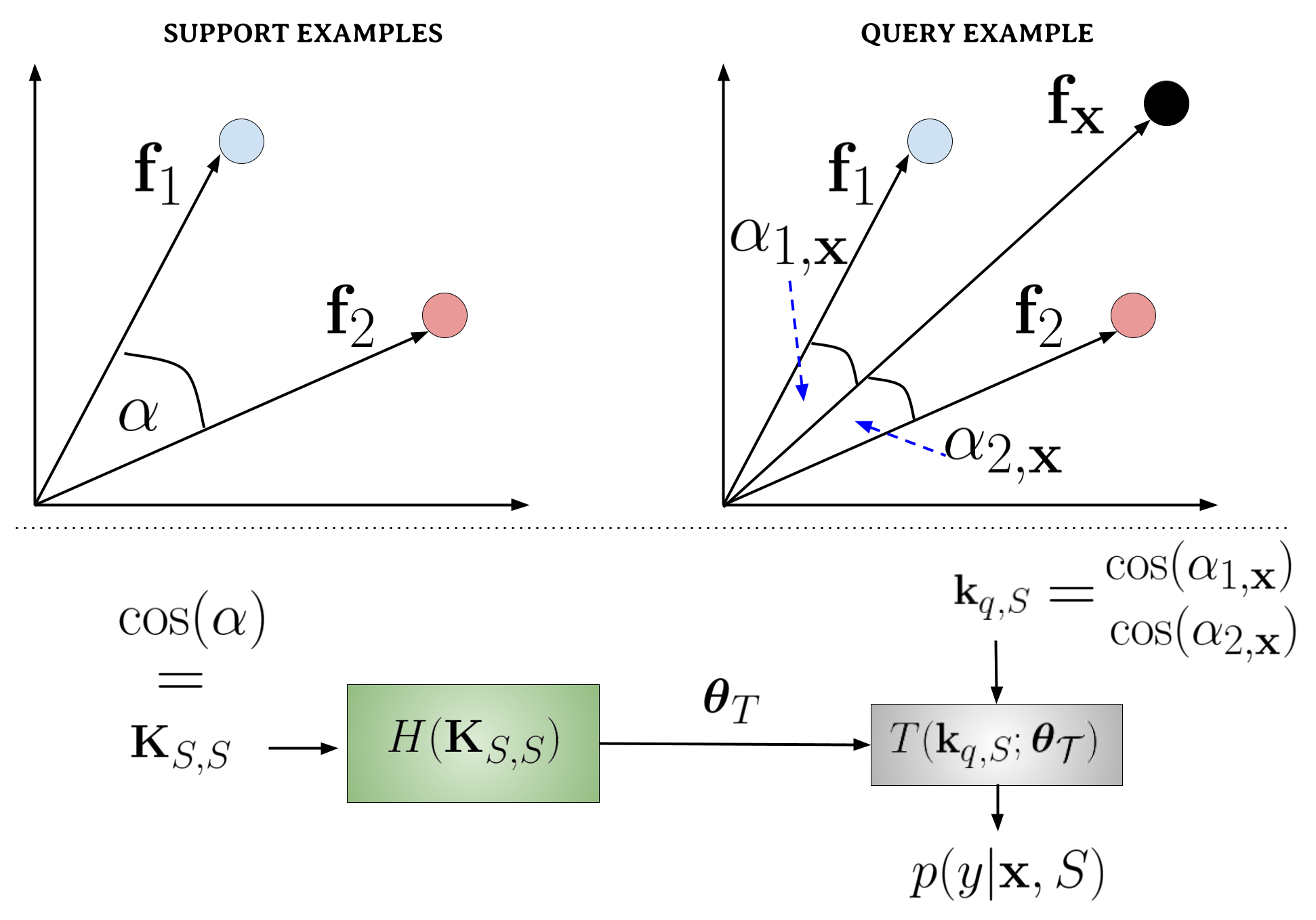}}
    \caption{Simple 2D example illustrating the application of cosine kernel for \our{}. We consider the two support examples from different classes represented by vectors $\mathbf{f_1}$ and $\mathbf{f_2}$. For this simple scenario, the input of hypernetwork is represented simply by the cosine of $\alpha$, which is an angle between vectors $\mathbf{f}_1$ and $\mathbf{f}_2$. We aim at classifying the query example $\mathbf{x}$ represented by a vector $\mathbf{f}_{\mathbf{x}}$. Considering our approach, we deliver to the target network $T(\cdot)$ the cosine values of angles between first ($\alpha_{\mathbf{x},1}$) and second ($\alpha_{\mathbf{x},2}$) support vectors and classify the query example using the weights $\boldsymbol{\theta}_{T}$ created by hypernetwork $H(\cdot)$ from $\cos{\alpha}$ (remaining components on the diagonal of $\mathbf{K}_{S,S}$ are constant for cosine kernel). }
    \label{fig:cosine}
    \end{center}
   \vskip -0.45in
\end{figure}

\subsection{Kernel function}

One of the key components of our approach is a kernel function $k(\cdot, \cdot)$. In this work we consider the dot product of the transformed vectors given by:

\begin{equation}
k(\mathbf{z}_1, \mathbf{z}_2)=\mathbf{f}(\mathbf{z}_1)^{\mathrm{T}} \mathbf{f}(\mathbf{z}_2),
\label{eq:kernel}
\end{equation}
where $\mathbf{f}(\cdot)$ can be a parametrized transformation function, represented by MLP model, or simply an identity operation, $\mathbf{f}(\mathbf{z})=\mathbf{z}$. In Euclidean space this criterion can be expressed as $k(\mathbf{z}_1, \mathbf{z}_2)=||\mathbf{f}(\mathbf{z}_1)|| \cdot||\mathbf{f}(\mathbf{z}_2)|| \cos{\alpha}$, where $\alpha$ is an angle between vectors $\mathbf{f}(\mathbf{z}_1)$ and $\mathbf{f}(\mathbf{z}_2)$. The main feature of this function is that it considers the vectors' norms, which can be problematic for some tasks that are outliers regarding the representations created by $\mathbf{f}(\cdot)$. Therefore, we consider in our experiments also the cosine kernel function given by:

\begin{equation}
k_{c}(\mathbf{z}_1, \mathbf{z}_2)=\frac{\mathbf{f}(\mathbf{z}_1)^{\mathrm{T}}\mathbf{f}(\mathbf{z}_2)}{||\mathbf{f}(\mathbf{z}_1)|| \cdot ||\mathbf{f}(\mathbf{z}_2)||} ,
\end{equation}
that represents the normalized version dot product. Considering the geometrical representation, $k_{c}(\mathbf{z}_1, \mathbf{z}_2)$ can be expressed as $\cos{\alpha}$ (see the example given by Fig. \ref{fig:cosine}). The support set is represented by two examples from different classes, $\mathbf{f}_1$ and $\mathbf{f}_2$. The target model parameters $\boldsymbol{\theta}_{T}$ are created based only on the cosine value of the angle between vectors $\mathbf{f}_1$ and $\mathbf{f}_2$. During the classification stage, the query example is represented by $\mathbf{f}_{\mathbf{x}}$, and the classification is applied on the cosine values of angles between $\mathbf{f}_{\mathbf{x}}$ and $\mathbf{f}_1$, and $\mathbf{f}_{\mathbf{x}}$ and $\mathbf{f}_2$, respectively.

\begin{algorithm}[t]
\small
    \caption{\our{} - training and prediction functions}
    \label{alg_overview}
    \textbf{Require:} Training set $\mathcal{D} = \{\mathcal{T}_n\}^N_{n=1}$, and $\mathcal{T}_{*}=\{\mathcal{S}_{*}, \mathcal{X}_{*}\}$ test task. \\
    \textbf{Parameters:}  $\boldsymbol{\theta}_H$ -  parameters , $\boldsymbol{\theta}_k$ - kernel parameters, and $\boldsymbol{\theta}_E$ - encoder parameters\\ 
    \textbf{Hyperparameters:} $N_{train}$ - number of training iterations, $N_{tune}$ number of tuning iterations, $\alpha$- step size.
    \begin{algorithmic}[1]
        \vspace{0.1cm} 
        \Function{Train}{$\mathcal{D}$, $\alpha$,  $N_{train}$, $\boldsymbol{\theta}_H$, $\boldsymbol{\theta}_k$, $\boldsymbol{\theta}_E$}
            \While{$n \leq N_{train}$}
                \State Sample task $\mathcal{T}=\{\mathcal{S}, \mathcal{Q}\} \sim \mathcal{D}$
                \State Assign support $\mathcal{S}=\{(\mathbf{x}_m,\mathbf{y}_m)\}_{m=1}^M$
                \State $L=-\sum_{m=1}^M \sum_{k=1}^K y_{m}^k \log p(y_{m}^k|\mathcal{S}_i, \mathbf{x}_{m},\boldsymbol{\theta}_H, \boldsymbol{\theta}_k, \boldsymbol{\theta}_E)$
                    \State Update: $\boldsymbol{\theta}_E \leftarrow \boldsymbol{\theta}_E - \alpha \nabla_{\theta_E} \mathcal{L}$,
                    \State \ \ \ \ \ \ \ \ \ \ \ \ \ \ $\boldsymbol{\theta}_H \leftarrow \boldsymbol{\theta}_H - \alpha \nabla_{\theta_H} \mathcal{L}$,
                    \State \ \ \ \ \ \ \ \ \ \ \ \ \ \ $\boldsymbol{\theta}_k \leftarrow \boldsymbol{\theta}_k - \alpha \nabla_{\theta_k} \mathcal{L}$
                    \State $n = n + 1$
            \EndWhile
            \State \Return $\boldsymbol{\theta}_H$, $\boldsymbol{\theta}_k$, $\boldsymbol{\theta}_E$
        \EndFunction
        \vspace{0.1cm} 
            \Function{Predict}{$\mathcal{T}_{*}$, $\alpha$, $N_{tune}$, $\boldsymbol{\theta}_H$, $\boldsymbol{\theta}_k$, $\boldsymbol{\theta}_E$}
            \State Create tuning task: $\mathcal{T}_t=\{\mathcal{S}_{*}, \mathcal{S}_{*}\}$
            \State Tune $\boldsymbol{\hat{\theta}}_H$, $\boldsymbol{\hat{\theta}}_k$, $\boldsymbol{\hat{\theta}}_E$ = \Call{Train}{$\mathcal{T}_t$, $\alpha$,  $N_{tune}$, $\boldsymbol{\theta}_H$, $\boldsymbol{\theta}_k$, $\boldsymbol{\theta}_E$}
            \For{each $\mathbf{x} \in \mathcal{X}_{*}$}
                \State \Return $\arg \max_{\mathbf{y}} p(\mathbf{y}|\mathcal{S}_{*}, \mathbf{x},\boldsymbol{\hat{\theta}}_H, \boldsymbol{\hat{\theta}}_k, \boldsymbol{\hat{\theta}}_E)$
            \EndFor
        \EndFunction
    \end{algorithmic}
\end{algorithm}

\subsection{Training and prediction}

The training procedure assumes the following parametrization of the model components. The encoder $E:=E_{\boldsymbol{\theta}_E}$ is parametrized by $\boldsymbol{\theta_E}$, the hypernetwork $H=H_{\boldsymbol{\theta}_H}$ by $\boldsymbol{\theta}_H$, and the kernel function $k$ by $\boldsymbol{\theta}_k$. We assume that training set $\mathcal{D}$ is represented by tasks $\mathcal{T}_i$ composed of support $\mathcal{S}_i$ and query $\mathcal{Q}_i$ examples. The training is performed by optimizing the cross-entropy criterion:

\begin{equation}
L = - \sum_{\mathcal{T}_i \in \mathcal{D}} \sum_{m=1}^M \sum_{k=1}^K y_{i, m}^k \log p(y_{i, m}^k|\mathcal{S}_i, \mathbf{x}_{i, m}),    
\end{equation}
where $(\mathbf{x}_{i,n}, \mathbf{y}_{i, n})$ are examples from query set $\mathcal{Q}_i$, where $Q_i=\{(\mathbf{x}_{i,m}, \mathbf{y}_{i, m})\}_{m=1}^M$. The distribution for currently considered classes $p(\mathbf{y}|\mathcal{S}, \mathbf{x})$ is returned by target network T of \our{}. During the training, we jointly optimize the parameters $\boldsymbol{\theta}_H$, $\boldsymbol{\theta}_k$,  and $\boldsymbol{\theta}_E$, minimizing the $L$ loss.  

During the inference stage, we consider the task $\mathcal{T}_*$, composed of a set of labeled support examples $\mathcal{S}_*$ and a set of unlabelled query examples represented by input values $\mathcal{X}_*$ that the model should classify. We can simply take the probability values $p(\mathbf{y}|\mathcal{S}_*, \mathbf{x})$ assuming the given support set $\mathcal{S}_*$ and single query observation $\mathbf{x}$ from $\mathcal{X}_*$, using the model with trained parameters $\boldsymbol{\theta}_H$, $\boldsymbol{\theta}_k$,  and $\boldsymbol{\theta}_E$. However, we observe that slightly better results are obtained while tuning the model's parameters on the considered task. We do not have access to  labels for query examples. Therefore we imitate the query set for this task simply by taking support examples and creating the tuning task $\mathcal{T}_i=\{\mathcal{S}_*,\mathcal{S}_*\}$ and updating the parameters of the model using several gradient iterations. The detailed presentation o training and prediction procedures are provided by Algorithm \ref{alg_overview}. 

\subsection{Adaptation to few-shot scenarios}

The proposed approach uses the ordering function $\pi(\cdot)$ that keeps the consistency between support kernel matrix $\mathbf{K}_{S,S}$ and the vector of kernel values $\mathbf{k}_{\mathbf{x}, S}$ for query example $\mathbf{x}$. For few-shot scenarios, each class has more than one representative in the support set. As a consequence, there are various possibilities to order the feature vectors in the support set inside the considered class. To eliminate this issue, we propose to apply the aggregation function to the embeddings $\mathbf{z}$ considering the support examples from the same class. Thanks to this approach, the kernel matrix is calculated based on the aggregated values of the latent space of encoding network $E$, making our approach independent of the ordering among the embeddings from the same class. In experimental studies, we examine the quality of \emph{mean} aggregation operation (\textbf{averaged}) against simple class-wise concatenation of the embeddings (\textbf{fine-grained}) in ablation studies.


\section{Related Work}

Feature transfer \cite{zhuang2020comprehensive} is a baseline procedure for few-shot learning and consists of pre-training the neural network and a classifier. During meta-validation, the classifier is fine-tuned to the novel tasks. \cite{chen2019closer} extend this idea by using cosine distance between the examples. 

In recent years, a variety of meta-learning methods \cite{hospedales2020metalearning,schmidhuber1992fast,bengio1992optimization} have been proposed to tackle the problem of few-shot learning. The various meta-learning architectures for few-shot learning can be roughly categorized into several groups: 

{\em Memory-based methods} \cite{ravi2016optimization,munkhdalai2018rapid,santoro2016meta,mishra2018simple,munkhdalai2017meta,zhen2020learning} are based on the idea to train a meta-learner with memory to learn novel concepts.

{\em Metric-based methods} meta-learn a deep representation with a metric in feature space, such that distance between examples from the support and query set with the same class have a small distance in such space. Some of the earliest works exploring this notion are Matching Networks \cite{vinyals2016matching} and Prototypical Networks \cite{snell2017prototypical}, which form \textit{prototypes} based on embeddings of the examples from the support set in the learned feature space and classify the query set based on the distance to those prototypes. Numerous subsequent works aim to improve the expressiveness of the prototypes through various techniques. \cite{oreshkin2018tadam} achieve this by conditioning the network on specific tasks, thus making the learned space task-dependent.  \cite{hu2021leveraging} transform embeddings of support and query examples in the feature space to make their distributions closer to Gaussian.
\cite{sung2018learning} propose Relation Nets, which learn the metric function instead of using a fixed one, such as Euclidean or cosine distance. 

Similar to the above methods, \our{} uses a kernel function that predicts the relations between the examples in a given task. The key difference is that instead of performing a nearest-neighbor classification based on the kernel values, in \our{}, the kernel matrix is classified by a task-specific classifier generated by the hypernetwork.

{\em Optimization-based methods} follow the idea of an optimization process over support set within the meta-learning framework like MetaOptNet \cite{lee2019meta}, Model-Agnostic Meta-Learning (MAML), and its extensions \cite{finn2017model,nichol2018first,raghu2019rapid,rajeswaran2019meta,finn2018probabilistic,nichol2018first}. Those techniques aim to train general models, which can adapt their parameters to the support set at hand in a small number of gradient steps. Similar to such techniques, \our{} also aims to produce task-specific models but utilizes a hypernetwork instead of optimization to achieve that goal.

{\em Gaussian processes} \cite{rasmussen2003gaussian} possess many properties useful in few-shot learning, such as natural robustness to the limited amounts of data and the ability to estimate uncertainty. When combined with meta-learned deep kernels,  \cite{patacchiola2020bayesian}, Gaussian processes were demonstrated to be a suitable tool for few-shot regression and classification, dubbed Deep Kernel Transfer (DKT). The assumption that such a universal deep kernel has enough data to generalize well to unseen tasks has been challenged in subsequent works. \cite{wang2021learning} introduced a technique of learning dense Gaussian processes by inducing variables. This approach achieves substantial performance improvement over the alternative methods. Similarly, \our{} also depends on learning a model that estimates task-specific functions' parameters. However, \our{} employs a hypernetwork instead of a Gaussian process to achieve that goal. 

{\em Hypernetworks} \cite{ha2016hypernetworks} have been proposed as a solution to few-shot learning problems in a number of works but have not been researched as widely as the approaches mentioned above. Multiple works proposed various variations of hyper-networks that predict a shallow classifier's parameters given the support examples \cite{bauer2017discriminative,qiao2017fewshot}. Subsequent works have extended those models by calculating cosine similarity between the query examples and the generated classifier weights \cite{gidaris2018dynamic} and utilizing a probabilistic model that predicts a distribution over the parameters suitable for the given task \cite{gordon2018meta}. More recently, \cite{zhmoginov2022hypertransformer} explored generating all of the parameters of the target network with a transformer-based hypernetwork, but found that for larger target networks, it is sufficient to generate only the parameters of the final classification layer.
A key characteristic of the above approaches is that during inference, the hypernetwork predicts weights responsible for classifying each class independently, based solely on the examples of that class from the support set. This property makes such solutions agnostic to the number of classes in a task, useful in practical applications. However, it also means that the hypernetwork does not take advantage of the inter-class differences in the task at hand.  

In contrast, \our{} exploits those differences by utilizing kernels, which helps improve its performance.


\section{Experiments}

In the typical few-shot learning setting, making a valuable and fair comparison between proposed models is often complicated because of the existence of the significant differences in architectures and implementations of known methods. In order to limit the influence of the deeper backbone (feature extractor) architectures, we follow the unified procedure proposed by \cite{chen2019closer}.

In this section, we describe the experimental analysis and performance of the \our{} in the large variety of few-shot benchmarks. Specifically, we consider both classification (see Section~\ref{sec:classification}) and cross-domain adaptation (see Section~\ref{sec:cross_domain}) tasks. Whereas the classification problems are focused on the most typical few-shot applications, the latter cross-domain benchmarks check the ability of the models to adapt to out-of-distribution tasks.
Additionally, we perform an ablation study of the possible adaptation procedures of \our{} to few-shot scenarios - presented in Section~\ref{sec:ablation}.

In all of the reported experiments, the tasks consist of 5 classes (5-way) and 1 or 5 support examples (1 or 5-shot). Unless indicated otherwise, all compared models use a known and widely utilized backbone consisting of four convolutional layers (each consisting of a 2D convolution, a batch-norm layer, and a ReLU non-linearity; each layer consists of 64 channels) and have been trained from scratch.

We report the performance of two variants of \our{}:
\begin{itemize}
    \item \textbf{\our{}} - models generated by the hypernetworks for each task.
    
    \item \textbf{\our{} + finetuning} - models generated by hypernetworks finetuned to the support examples of each task for 10 training steps\footnote{In the case of the finetuned hypernetworks, we tune a copy of the hypernetwork separately for each validation task. This way, we ensure that our model does not take unfair advantage of the validation tasks.}. 
\end{itemize}

In all cases, we observe a modest performance boost thanks to finetuning the hypernetwork.

Comprehensive details for each training procedure are reported in the Appendix.

\subsection{Classification}\label{sec:classification}

\begin{table}[t!]
\centering
\caption{The classification accuracy results for the inference tasks on $\textbf{CUB}$ and $\textbf{mini-ImageNet}$ datasets in the $1$-shot setting. The highest results are bold and second-highest in italic (the larger, the better).}
\label{tab:conv41shotminiimagenet}
\scalebox{0.65}{
\begin{tabular}{l@{\hspace*{5mm}}cc}
\toprule
\textbf{Method} & \textbf{CUB} & \textbf{mini-ImageNet}\\
\midrule
\textbf{ML-LSTM} \cite{ravi2016optimization} & --  & $43.44 \pm 0.77$  \\
\textbf{SNAIL} \cite{mishra2018simple} & -- & $45.10$  \\
\textbf{iMAML-HF} \cite{rajeswaran2019meta} & --  & $49.30 \pm 1.88$  \\
\textbf{LLAMA} \cite{grant2018recasting} & --  & $49.40 \pm 1.83$  \\
\textbf{VERSA} \cite{gordon2018meta} & -- & $48.53 \pm 1.84$  \\
\textbf{Amortized VI} \cite{gordon2018meta} & -- & $44.13 \pm 1.78$  \\
\textbf{Meta-Mixture} \cite{jerfel2019reconciling} & -- & $49.60 \pm 1.50$  \\
\textbf{SimpleShot} \cite{wang2019simpleshot} & -- & $49.69 \pm 0.19$  \\
\textbf{Feature Transfer} \cite{zhuang2020comprehensive}  & $46.19 \pm 0.64$ & $39.51 \pm 0.23$  \\
\textbf{Baseline++} \cite{chen2019closer} & $61.75 \pm 0.95$ & $47.15 \pm 0.49$  \\
\textbf{MatchingNet} \cite{vinyals2016matching} & $60.19 \pm 1.02$ & $48.25 \pm 0.65$  \\
\textbf{ProtoNet} \cite{snell2017prototypical} &  $52.52 \pm 1.90$ & $44.19 \pm 1.30$  \\
\textbf{MAML} \cite{finn2017model} & $56.11 \pm 0.69$ & $45.39 \pm 0.49$  \\
\textbf{RelationNet} \cite{sung2018learning} & $62.52 \pm 0.34$ & $48.76 \pm 0.17$  \\
\textbf{DKT + CosSim} \cite{patacchiola2020bayesian} & $63.37 \pm 0.19$ & $48.64 \pm 0.45$  \\
\textbf{DKT + BNCosSim} \cite{patacchiola2020bayesian} & $62.96 \pm 0.62$  & $49.73 \pm 0.07$  \\
\textbf{GPLDLA} \cite{kim2021gaussian} &  $63.40 \pm 0.14$ & $52.58 \pm 0.19$  \\
\makecell[cl]{\textbf{amortized Bayesian} \\  \textbf{prototype meta-learning}  \cite{sun2021amortized}}  & $63.46 \pm 0.98$ & $53.28 \pm 0.91$  \\
\textbf{VAMPIRE} \cite{nguyen2020uncertainty}& -- & $51.54 \pm 0.74$  \\
\textbf{PLATIPUS} \cite{finn2018probabilistic} & -- & $50.13 \pm 1.86$  \\
\textbf{ABML} \cite{ravi2018amortized} & $49.57 \pm 0.42$ & $45.00 \pm 0.60$  \\
\textbf{Bayesian MAML} \cite{yoon2018bayesian}  &  $55.93 \pm 0.71$ & $53.80 \pm 1.46$  \\
\textbf{OVE PG GP + Cosine (ML)} \cite{snell2020bayesian}   & $63.98 \pm 0.43$ & $50.02 \pm 0.35$ \\
\textbf{OVE PG GP + Cosine (PL)}  \cite{snell2020bayesian}  & $60.11 \pm 0.26$ &  $48.00 \pm 0.24$ \\
\textbf{Reptile} \cite{nichol2018first} & -- & $49.97 \pm 0.32$  \\
\textbf{R2-D2} \cite{bertinetto2018meta} & --  & $48.70 \pm 0.60$  \\
\textbf{VSM} \cite{zhen2020learning} & --  & $\mathit{54.73 \pm 1.60}$  \\
\textbf{PPA} \cite{qiao2017fewshot} & --  & $54.53 \pm 0.40$ \\
\textbf{DFSVLwF} \cite{gidaris2018dynamic} & -- & $\mathbf{56.20 \pm 0.86}$ \\
\midrule
\textbf{\our{}} (ours)  & $\mathit{65.27 \pm 0.24}$  & $52.42 \pm 0.46$  \\
\textbf{\our{} + finetuning} (ours) & $\mathbf{66.13 \pm 0.26}$   & $53.18 \pm 0.45$  \\
\bottomrule
\end{tabular}
}
\end{table}

\begin{table}[t!]
\centering
\caption{The classification accuracy results for the inference tasks on $\textbf{CUB}$ and $\textbf{mini-ImageNet}$ datasets in the $5$-shot setting. The highest results are bold and second-highest in italic (the larger, the better). }
\label{tab:conv45shotcubminiimagenet}
\scalebox{0.65}{
\begin{tabular}{l@{\hspace*{5mm}}cc}
\toprule
\textbf{Method}    & \textbf{CUB} & \textbf{mini-ImageNet} \\ 
\midrule
\textbf{ML-LSTM} \cite{ravi2016optimization} & -- & $60.60 \pm 0.71$ \\
\textbf{SNAIL} \cite{mishra2018simple} & -- & $55.20$ \\
\textbf{VERSA} \cite{gordon2018meta}   & -- & $67.37 \pm 0.86$ \\
\textbf{Amortized VI} \cite{gordon2018meta} & -- & $55.68 \pm 0.91$ \\
\textbf{Meta-Mixture} \cite{jerfel2019reconciling} & -- & $64.60 \pm 0.92$ \\
\textbf{SimpleShot} \cite{wang2019simpleshot} & -- & $66.92 \pm 0.17$ \\
\textbf{Feature Transfer}    & $68.40 \pm 0.79$ & $60.51 \pm 0.55$ \\
\textbf{Baseline++}  \cite{chen2019closer}  & $78.51 \pm 0.59$ & $66.18 \pm 0.18$ \\
\textbf{MatchingNet} \cite{vinyals2016matching}   & $75.11 \pm 0.35$ & $62.71 \pm 0.44$ \\
\textbf{ProtoNet} \cite{snell2017prototypical}   & $75.93 \pm 0.46$ & $64.07 \pm 0.65$ \\
\textbf{MAML} \cite{finn2017model}   & $74.84 \pm 0.62$ & $61.58 \pm 0.53$ \\
\textbf{RelationNet}  \cite{sung2018learning}  & $78.22 \pm 0.07$ & $64.20 \pm 0.28$ \\
\textbf{DKT + CosSim} \cite{patacchiola2020bayesian}   & $77.73 \pm 0.26$ & $62.85 \pm 0.37$ \\
\textbf{DKT + BNCosSim} \cite{patacchiola2020bayesian}   & $77.76 \pm 0.62$ & $64.00 \pm 0.09$ \\
\textbf{GPLDLA} \cite{kim2021gaussian} & $78.86 \pm 0.35$ & -- \\
\makecell[cl]{\textbf{amortized Bayesian} \\ \textbf{prototype meta-learning} \cite{sun2021amortized}}  &  $\mathbf{80.94 \pm 0.62}$ & $\mathbf{70.44 \pm 0.72}$ \\
\textbf{VAMPIRE} \cite{nguyen2020uncertainty}   & -- & $64.31 \pm 0.74$ \\
\textbf{ABML} \cite{ravi2018amortized} & $68.94 \pm 0.16$ & -- \\
\textbf{Bayesian MAML} \cite{yoon2018bayesian} & -- & $64.23 \pm 0.69$ \\
\textbf{OVE PG GP + Cosine (ML)} \cite{snell2020bayesian}   & $77.44 \pm 0.18$ & $64.58 \pm 0.31$\\
\textbf{OVE PG GP + Cosine (PL)}  \cite{snell2020bayesian}  & $79.07 \pm 0.05$ & $67.14 \pm 0.23$ \\
\textbf{Reptile} \cite{nichol2018first} & -- & $65.99 \pm 0.58$  \\
\textbf{R2-D2} \cite{bertinetto2018meta} & -- & $65.50 \pm 0.60$  \\
\textbf{VSM} \cite{zhen2020learning} & -- & $68.01 \pm 0.90$  \\
\midrule
\textbf{\our{}}   & $ 79.80 \pm 0.16$ & $68.78 \pm 0.29$ \\
\textbf{\our{} + finetuning}   & $\mathit{80.07 \pm 0.22}$ & $\mathit{69.62 \pm 0.28}$ \\
\bottomrule
\end{tabular}
}
\end{table}

Firstly, we consider a classical few-shot learning scenario, where all the classification tasks (both training and inference) come from the same dataset. The main aim of the proposed classification experiments is to find the ability of the few-shot models to adapt to never-seen tasks from the same data distribution. 

We benchmark the performance of the \our{} and other methods on two challenging and widely considered datasets: Caltech-USCD Birds (\textbf{CUB}) \cite{wah2011cub} and \textbf{mini-ImageNet} \cite{ravi2016optimization}. The following experiments are in the most popular setting, 5-way, consisting of 5 random classes. In all experiments, the query set of each task consists of 16 samples for each class (80 in total).  We provide the additional training details in the Appendix.
We compare \our{} to a vast pool of the state-of-the-art algorithms, including the canonical methods (like Matching Networks \cite{vinyals2016matching}, Prototypical Networks \cite{snell2017prototypical}, MAML \cite{finn2017model}, and its extensions) as well as the recently popular Bayesian methods mostly build upon the Gaussian Processes framework (like DKT \cite{patacchiola2020bayesian}).

We first consider the more challenging 1-shot task and report the results in Table \ref{tab:conv41shotminiimagenet}. We further consider the 5-shot setting and report the results in Table \ref{tab:conv45shotcubminiimagenet}. The additional results comparing methods on larger backbones are included in Appendix.

In the $1$-shot scenario, \our{} achieves the highest accuracies in the \textbf{CUB} dataset with and without utilizing a finetuning procedure ($66.13\%$ with a finetuning, $65.27\%$ without) and performs better than any other model. However, in the \textbf{mini-ImageNet} dataset, our approach is among the best models ($53.18\%$), slightly losing with DFSVLwF \cite{gidaris2018dynamic} ($56.20\%$).

Considering the $5$-shot scenario, \our{} is the second-best model achieving $80.07\%$ in the \textbf{CUB} dataset and $69.62\%$ in the \textbf{mini-ImageNet}, whereas the best model, amortized Bayesian prototype meta-learning, is insignificantly better and achieves $80.94\%$ and $70.44\%$ on the mentioned datasets, respectively. 

The obtained results clearly show that \our{} achieves state-of-the-art or comparable results to the best models on the intensive set of a standard classification few-shot learning setting.

\subsection{Cross-domain adaptation}\label{sec:cross_domain}

\begin{table*}[t!]
\centering
\caption{The classification accuracy results for the inference tasks on cross-domain tasks (\textbf{Omniglot}$\rightarrow$\textbf{EMNIST} and \textbf{mini-ImageNet}$\rightarrow$\textbf{CUB}) datasets in the $1$-shot setting. The highest results are bold and second-highest in italic (the larger, the better). }
\scalebox{0.90}{
\begin{tabular}{lcccc}
\hline
\textbf{} & \multicolumn{2}{c}{\textbf{Omni}$\rightarrow$\textbf{EMNIST}} & \multicolumn{2}{c}{\textbf{mini-ImageNet}$\rightarrow$\textbf{CUB}} \\
 {\textbf{Method}} & \textbf{1-shot}& \textbf{5-shot} & \textbf{1-shot} & \textbf{5-shot} \\ 
\hline
 {\textbf{Feature Transfer}} & 64.22 $\pm$  {1.24} & 86.10 $\pm$  {0.84} & 32.77 $\pm$  {0.35} & 50.34 $\pm$  {0.27}\\
 {\textbf{Baseline$++$}} \cite{chen2019closer} & 56.84 $\pm$  {0.91} & 80.01 $\pm$  {0.92} & 39.19 $\pm$  {0.12} & $ 57.31 \pm 0.11 $ \\
 {\textbf{MatchingNet}} \cite{vinyals2016matching}  & 75.01 $\pm$  {2.09} & 87.41 $\pm$  {1.79} & 36.98 $\pm$  {0.06} & 50.72 $\pm$  {0.36} \\
 {\textbf{ProtoNet}} \cite{snell2017prototypical} & 72.04 $\pm$  {0.82} & 87.22 $\pm$  {1.01} & 33.27 $\pm$  {1.09} & 52.16 $\pm$  {0.17} \\
 {\textbf{MAML}} \cite{finn2017model} & 72.68 $\pm$  {1.85} & 83.54 $\pm$  {1.79}  & 34.01 $\pm$  {1.25} &48.83 $\pm$  {0.62} \\
 {\textbf{RelationNet}} \cite{sung2018learning} & 75.62 $\pm$  {1.00} & 87.84 $\pm$  {0.27}  & 37.13 $\pm$  {0.20} & 51.76 $\pm$  {1.48}\\
 {\textbf{DKT}} \cite{patacchiola2020bayesian} & 75.40 $\pm$  {1.10} & $\mathit{90.30 \pm  0.49}$ & $\mathbf{40.14 \pm  {0.18}}$ & 56.40 $\pm$  {1.34} \\
\textbf{Bayesian MAML} \cite{yoon2018bayesian} & $63.94 \pm 0.47$ & $65.26 \pm 0.30 $ & $33.52 \pm 0.36 $ & $51.35 \pm 0.16$ \\

\textbf{OVE PG GP + Cosine (ML)} \cite{snell2020bayesian} & $68.43 \pm 0.67 $ & $86.22 \pm 0.20 $ & $39.66 \pm 0.18 $ & $55.71 \pm 0.31$ \\
\textbf{OVE PG GP + Cosine (PL)}  \cite{snell2020bayesian}  & $ 77.00 \pm 0.50 $ & $87.52 \pm 0.19 $ & $37.49 \pm 0.11$ & $57.23 \pm 0.31$ \\
\hline 
 \textbf{\our{} } &  $\mathit{78.06 \pm 0.24}$ & $89.04 \pm 0.18 $ & $39.09 \pm 0.28$ & $\mathit{57.77 \pm 0.33}$ \\
 \textbf{\our{} + finetuning} & $\mathbf{80.65 \pm 0.30}$ & $\mathbf{90.81 \pm 0.16}$ & $\mathit{40.03 \pm 0.41 }$ & $\mathbf{58.86 \pm 0.38}$ \\

\hline
\end{tabular}
}
\label{tab:crossdomain_accuracy}
\end{table*}

In the cross-domain adaptation setting, the model is evaluated on tasks coming from a different distribution than the one it had been trained on. Therefore, such a task is more challenging than standard classification and is a plausible indicator of a model’s ability to generalize. In order to benchmark the performance of \our{} in cross-domain adaptation, we merge data from two datasets so that the training fold is drawn from the first dataset and validation and testing fold -- from another one. Specifically, we test \our{} on two cross-domain classification tasks: \\
\textbf{mini-ImageNet $\rightarrow$ CUB} (model trained on \textbf{mini-ImageNet} and evaluated on \textbf{CUB}) and \textbf{Omniglot $\rightarrow$ EMNIST} in the $1$-shot and $5$-shot settings. We report the results in Table \ref{tab:crossdomain_accuracy}. In most settings, \our{} achieves the highest accuracy, except for $1$-shot \textbf{mini-ImageNet $\rightarrow$ CUB} classification, where its accuracy is on par with the accuracy achieved by DKT \cite{patacchiola2020bayesian} ($40.14\%$ and $40.03\%$ achieved by DKT and \our{}, respectively). We note that just in the case of regular classification, finetuning the hypernetwork on the individual tasks consistently improves its performance.

\subsection{Aggregating support examples in the 5-shot setting}\label{sec:ablation}

In \our{}, the hypernetwork generates the weights of the information about the support examples, expressed through the support-support kernel matrix. In the case of $5$-way $1$-shot classification, each task consists of $5$ support examples, and therefore, the size of the kernel matrix is $(5 \times 5)$, and the input size of the hypernetwork is $25$. However, when the number of support examples grows, increasing the size of the kernel matrix may be impractical and lead to overparametrization of the hypernetwork. 

Since hypernetworks are known to be sensitive to large input sizes \cite{ha2016hypernetworks}, we consider a way to maintain a constant input size of \our{}, independent of the number of support examples of each class by using means of support embeddings of each class for kernel calculation, instead of individual embeddings. Prior works suggest that when there are multiple examples of a class, an averaged embedding of such class represents it sufficiently in the embedding space \cite{snell2017prototypical}. 

To verify this approach, in the $5$-shot setting, we train \our{} with two variants of calculating the inputs to the kernel matrix:

\begin{itemize}
    \item \textbf{fine-grained} -- utilizing a hypernetwork that takes as an input a kernel matrix between each of the embeddings of the individual support examples. This kernel matrix has a shape of $(25 \times 25)$.
    \item \textbf{averaged} -- utilizing a hypernetwork where the kernel matrix is calculated between the \textbf{means} of embeddings of each class. The kernel matrix in this approach has a shape of $(5 \times 5)$.
\end{itemize}
                                
We benchmark both variants of \our{} on the $5$-shot classification task on \textbf{CUB} and \textbf{mini-ImageNet} datasets. Moreover, we compare these approaches also on cross-domain classification between the Omniglot and EMNIST datasets. We report the accuracies in Table \ref{tab:aggregation_ablation}. It is evident that averaging the embeddings before calculating the kernel matrix yields superior results.

\begin{table}[h!]
\caption{The classification \textit{accuracy} results for \our{} in the $5$-shot setting with two variants of the support embeddings aggregation. The performance measured on \textbf{Omniglot}$\rightarrow$\textbf{EMNIST}, $\textbf{CUB}$, and \textbf{mini-ImageNet}$\rightarrow$\textbf{CUB} tasks. The larger, the better.}
\label{tab:aggregation_ablation}
\centering
\scalebox{0.67}{
\begin{tabular}{lcccc}
\hline
\textbf{} & \textbf{Omni}$\rightarrow$\textbf{EMNIST} & \textbf{CUB} & \textbf{mini-ImageNet} \\
\hline
 \textbf{\our{} (fine-grained)} &  $87.55 \pm 0.19$ & $78.05 \pm 0.20$ & $67.07 \pm 0.47$  \\
 \textbf{\our{} (averaged)} & $89.04 \pm 0.18 $ & $79.80 \pm 0.16$ & $69.62 \pm 0.28$  \\
\hline
\end{tabular}
}
\end{table}

\section{Conclusion}
In this work, we introduced \our{} –- a new framework that uses kernel methods combined with hypernetworks. Our method uses the kernel-based representation of the support examples and a hypernetwork paradigm to create the query set's classification module. We concentrate on relations between embeddings of the support examples instead of direct feature values. Thanks to this approach, our model can adapt to highly different tasks.

We evaluate the \our{} model on various one-shot and few-shot image classification tasks. \our{} demonstrates high accuracy in all tasks, performing comparably or better to state-of-the-art solutions. Furthermore, the model has a strong ability to generalize, as evidenced by its performance on cross-domain classification tasks.

\bibliography{ref}
\bibliographystyle{icml2022}

\newpage
\appendix
\onecolumn
\section{Additional results - ResNet10}
\label{app:resnet}

This section provides additional results in the $5$-way ($1$-shot and $5$-shot) classification tasks for models using a larger backbone, namely ResNet-10 \cite{he2015resnet}. We provide the results for \textbf{CUB} and \textbf{mini-ImageNet} datasets in Tables \ref{tab:cubresnet} and \ref{tab:miniresnet}.

In the \textbf{CUB} dataset classification tasks (see Table \ref{tab:cubresnet}), \our{} is amongst the state-of-the-art models achieving classification accuracy often equal within the variance to the best models. Considering the $5$-shot scenario, the highest classification result across the evaluated methods ($86.38\% \pm 0.15$) obtained the GPLDLA model based on the Gaussian Processes framework. However, the \our{} performance, $86.28\% \pm 0.29$, is the second-best but even lies within the variance of the best model. In the $1$-shot setting, ProtoNet obtains the highest result ($73.22\% \pm 0.92$), whereas \our{} is the third one ($71.99\% \pm 0.70$) but still equal according to the variances.

In the \textbf{mini-ImageNet} classification task, \our{} achieves the second-best accuracy in both 1-shot and 5-shot settings\footnote{In the case of the \textbf{mini-ImageNet} classification with ResNet10, we benchmarked all of the listed models ourselves. To our best knowledge, previously, there were no reported benchmarks on this dataset with the ResNet-10 backbone.}. In the $1$-shot setting, the DKT model \cite{patacchiola2020bayesian} achieved the best result, with \our{} being a close second, with only $0.04$ pp difference. In the $5$-shot setting, the baseline++ approach outperforms all others by a large margin \cite{chen2019closer}, whereas \our{} and ProtoNet \cite{snell2017prototypical} achieve similar, second-best results. We observe that apart from \our{}, which achieves second-best results in both settings, models which perform well in one setting are outperformed by others in the second and vice versa.

It is worth noticing that \our{} without finetuning steps performances sometimes slightly better than the same with finetuning. We even observe that a few first steps of finetuning procedure result in an unnoticeable increase of accuracy of the basic model. However, the usual $10$ steps result in this setting in slightly worse performance, so one should use it cautiously. We decided to report the results after the standard finetuning procedure only.

\begin{table*}[h!]
\centering
\caption{ The classification accuracy results for the inference tasks in the \textbf{CUB} dataset in the 5-way (1-shot and 5-shot) scenarios. We consider models using the ResNet-10 backbone. The highest results are bold and second-highest in italic (the larger, the better).}
\label{tab:cubresnet}
\begin{tabular}{lcc}
\toprule
\textbf{Method}    & \textbf{1-shot} & \textbf{5-shot} \\
\midrule
\textbf{Feature Transfer}    &  $63.64 \pm 0.91$ & $81.27 \pm 0.57$ \\
\textbf{Baseline++}  \cite{chen2019closer}  & $69.55 \pm 0.89$ & $85.17 \pm 0.50$  \\
\textbf{MatchingNet} \cite{vinyals2016matching}   & $71.29 \pm 0.87$ & $83.47 \pm 0.58$  \\
\textbf{ProtoNet} \cite{snell2017prototypical}   & $\mathbf{73.22 \pm 0.92}$ & $85.01 \pm 0.52$ \\
\textbf{MAML} \cite{finn2017model}   & $70.32 \pm 0.99$ & $80.93 \pm 0.71$ \\
\textbf{RelationNet}  \cite{sung2018learning}  &  $70.47  \pm 0.99 $ & $83.70 \pm 0.55$ \\
\textbf{DKT + CosSim} \cite{patacchiola2020bayesian}   & $70.81 \pm 0.52$ & $83.26 \pm 0.50$ \\
\textbf{DKT + BNCosSim} \cite{patacchiola2020bayesian}   & $\mathit{72.27 \pm 0.30}$ & $85.64 \pm 0.29$ \\
\textbf{SimpleShot} \cite{wang2019simpleshot} & $53.78 \pm 0.21$ & $71.41 \pm 0.17$ \\
\textbf{GPLDLA} \cite{kim2021gaussian} & $71.30 \pm 0.16$ & $\mathbf{86.38 \pm 0.15}$ \\
\midrule
\textbf{\our{}}   & $71.99 \pm 0.70$ & $86.28 \pm 0.29$ \\
\textbf{\our{} + finetuning}   & $71.60 \pm 0.59$ & $\mathit{86.22 \pm 0.30}$ \\
\bottomrule
\end{tabular}
\end{table*}

\begin{table*}[h!]
\centering
\caption{
    The classification accuracy results for the inference tasks in the \textbf{mini-ImageNet} dataset in the 5-way (1-shot and 5-shot) scenarios. We consider models using the ResNet-10 backbone. The highest results are bold and second-highest in italic (the larger, the better).}
\label{tab:miniresnet}
\begin{tabular}{lcc}
\toprule
\textbf{Method}    & \textbf{1-shot} & \textbf{5-shot} \\
\midrule
\textbf{Baseline++}  \cite{chen2019closer}  & $54.35 \pm 0.34$ & $\mathbf{75.26 \pm 0.16}$ \\
\textbf{MatchingNet} \cite{vinyals2016matching}   & $54.18 \pm 0.09$ & $67.71 \pm 0.20$ \\
\textbf{ProtoNet} \cite{snell2017prototypical}   & $53.28 \pm 0.17$ & $73.04 \pm 0.15$ \\
\textbf{RelationNet}  \cite{sung2018learning}  & $51.88 \pm 0.45$  & $67.21 \pm 0.16$ \\
\textbf{DKT + BNCosSim} \cite{patacchiola2020bayesian}   & $\mathbf{56.03 \pm 0.50}$ & $71.28 \pm 0.12$ \\
\midrule
\textbf{\our{}} & $55.36 \pm 0.64$ & $\mathit{73.06 \pm 0.30}$  \\
\textbf{\our{} + finetuning}   & $\mathit{55.99 \pm 0.63}$ &   $72.87 \pm 0.33 $ \\
\bottomrule
\end{tabular}
\end{table*}

\section{Training details}

In this section, we present in detail the architecture and hyperparameters of \our{}.

\paragraph{Architecture overview}

    From a high-level perspective, the architecture of \our{} consists of three parts:
    \begin{itemize}
        \item backbone - a convolutional feature extractor.
        \item neck - a sequence of zero or more fully-connected layers with ReLU nonlinearities in between.
        \item heads - for each parameter of the target network, a sequence of one or more linear layers, which predicts the values of that parameter. All heads of \our{} have identical lengths, hidden sizes, and input sizes that depend on the generated parameter's size.
    \end{itemize}
    
\begin{figure}
        \centering
        \includegraphics[width=0.95\textwidth]{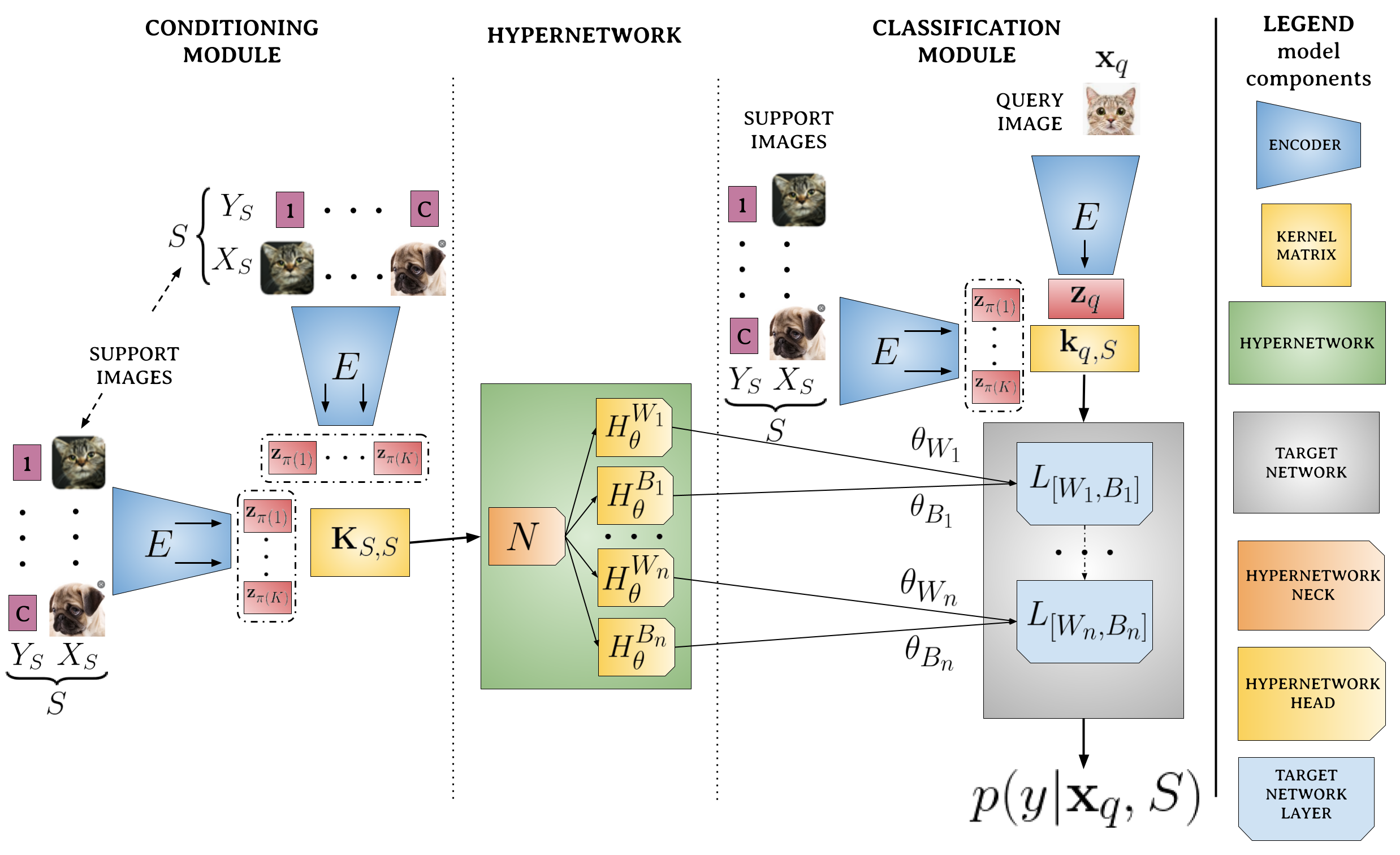}
        \caption{A detailed outline of the architecture of \our{}, with the denoted flow of parameters generated by the hypernetwork heads.
        }
        \label{fig:schema_flow}
    \end{figure}

    The target network generated by \our{} re-uses its backbone. We outline this architecture in Figure \ref{fig:schema_flow}.
    
\paragraph{Backbone}
    For each experiment described in the main body of this work, we follow  \cite{patacchiola2020bayesian} in using a shallow backbone (feature extractor) for \our{} as well as referential models. This backbone consists of four convolutional layers, each consisting of a convolution, batch normalization, and ReLU nonlinearity. Apart from the first convolution, which has the number of input size equal to the number of image channels, each convolution has an input and output size of 64. We apply max-pooling between each convolution, which decreases by half the resolution of the processed feature maps. The output of the backbone is flattened so that the further layers can process it.
    
    We perform additional experiments described in Appendix \ref{app:resnet} where instead of the above backbone, we utilize ResNet-10 \cite{he2015resnet}.
    
\paragraph{Datasets}
    For the purpose of making a fair comparison, we follow the procedure presented in, e.g., \cite{patacchiola2020bayesian,chen2019closer}. In the case of the \textbf{CUB} dataset \cite{wah2011cub}, we split the whole amount of $200$ classes ($11788$ images) across train, validation, and test consisting of $100$, $50$, and $50$ classes, respectively \cite{chen2019closer}. The \textbf{mini-ImageNet} dataset \cite{ravi2016optimization} is created as the subset of \textbf{ImageNet} \cite{russakovsky2015imagenet}, which consists of $100$ different classes represented by $600$ images for each one. We followed the standard procedure and divided the \textbf{mini-ImageNet} into $64$ classes for the train, $16$ for the validation set, and the remaining $20$ classes for the test. The well-known \textbf{Omniglot} dataset \cite{lake2011one} is a collection of characters from $50$ different languages. The \textbf{Omniglot} contains $1623$ white and black characters in total. We utilize the standard procedure to include the examples rotated by $90^\circ$ and increase the size of the dataset to $6492$, from which $4114$ were further used in training. Finally, the \textbf{EMNIST} dataset \cite{cohen2017emnist} collects the characters and digits coming from the English alphabet, which we split into $31$ classes for the test and $31$ for validation.
    
\paragraph{Data augmentation}
    We apply data augmentation during model training in all experiments, except \textbf{Omniglot} $\rightarrow$ \textbf{EMNIST} cross-domain classification. The augmentation pipeline is identical to the one used by \cite{patacchiola2020bayesian} and consists of the random crop, horizontal flip, and color jitter steps.

\subsection{Hyperparameters}

    Below, we outline the hyperparameters of architecture and training procedures used in each experiment.
    
    We use cosine similarity as a kernel function and averaged support embeddings aggregation in all experiments. \our{} is trained with the learning rate of $0.001$ with the Adam optimizer \cite{kingma2014adam} and no learning rate scheduler. Task-specific finetuning is also performed with the Adam optimizer and the learning rate of $0.0001$.
    
    For the natural image tasks (\textbf{CUB}, \textbf{mini-ImageNet}, \textbf{mini-ImageNet} $\rightarrow$ \textbf{CUB} classification), we use a hypernetwork with the neck length o $2$, head lengths of $3$, and a hidden size of $4096$, which produce a target network with a single fully-connected layer. We perform training for $10000$ epochs.
    
    For the simpler \textbf{Omniglot} $\rightarrow$ \textbf{EMNIST} character classification task, we train a smaller hypernetwork with the neck length of $1$, head lengths of $2$, and the hidden size of $512$, which produces a target network with two fully-connected layers and a hidden size of $128$. We train this hypernetwork for a shorter number of epochs, namely $2000$.
    
    We summarize all the above hyperparameters in Table \ref{tab:hyperparameters}.

\begin{table*}[t!]
\centering
\caption{Hyperparameters}
\label{tab:hyperparameters}
\scalebox{0.95}{
\begin{tabular}{lcccc}
\toprule
\textbf{hyperparameter}    & \textbf{CUB} & \textbf{mini-ImageNet} & \textbf{mini-ImageNet} $\rightarrow$ \textbf{CUB} & \textbf{Omniglot} $\rightarrow$ \textbf{EMNIST} \\
\midrule
kernel function & cosine similarity & cosine similarity & cosine similarity & cosine similarity \\
learning rate & $0.001$  & $0.001$ & $0.001$ & $0.001$ \\
hypernetwork's head layers no. & $3$  & $3$ & $3$ & $2$ \\
hypernetwork's neck layers no. & $2$  & $2$ & $2$ & $1$ \\
hypernetwork layers' hidden dim & $4096$  & $4096$ & $4096$ & $512$ \\
support embeddings aggregation & averaged  & averaged  & averaged  & averaged \\
taskset size & $1$  & $1$ & $1$& $1$ \\
target network layers no. & $1$  & $1$ & $1$ & $2$ \\
target network activation & ReLU  & ReLU & ReLU  & ReLU \\
finetuning epochs (if used) & $10$  & $10$ & $10$ & $10$ \\
finetuning learning rate & $0.0001$  & $0.0001$ & $0.0001$ & $0.0001$ \\
optimizer & Adam  & Adam & Adam & Adam\\
epochs no. & $10000$  & $10000$ & $10000$ & $2000$ \\
\bottomrule
\end{tabular}
}
\end{table*}

\end{document}